\documentclass{article}

\usepackage{microtype}
\usepackage{graphicx}
\usepackage{booktabs} 
\usepackage{pifont}
\usepackage{makecell}
\usepackage{subcaption}
\usepackage{amsmath}
\usepackage{multirow}
\usepackage{multicol}
\usepackage{stmaryrd}

\usepackage{hyperref}

\usepackage[capitalize]{cleveref}
\crefname{section}{Sec.}{Secs.}
\Crefname{section}{Section}{Sections}
\Crefname{table}{Table}{Tables}
\crefname{table}{Tab.}{Tabs.}

\usepackage{tikz}
\newcommand*\circled[1]{\tikz[baseline=(char.base)]{
            \node[shape=circle,draw,inner sep=1pt] (char) {#1};}}

\usepackage{amsmath,amsfonts,bm}

\def\eqref#1{equation~\ref{#1}}

\def\1{\bm{1}}

\def\rvx{{\mathbf{x}}}

\def\vtheta{{\bm{\theta}}}
\def\va{{\bm{a}}}

\def\vk{{\bm{k}}}

\def\vo{{\bm{o}}}

\def\vq{{\bm{q}}}

\def\vu{{\bm{u}}}
\def\vv{{\bm{v}}}
\def\vw{{\bm{w}}}

\def\mA{{\bm{A}}}

\def\mW{{\bm{W}}}

\DeclareMathAlphabet{\mathsfit}{\encodingdefault}{\sfdefault}{m}{sl}
\SetMathAlphabet{\mathsfit}{bold}{\encodingdefault}{\sfdefault}{bx}{n}

\def\sA{{\mathbb{A}}}

\def\sH{{\mathbb{H}}}
\def\sI{{\mathbb{I}}}

\def\sR{{\mathbb{R}}}

\def\sX{{\mathbb{X}}}

\newcommand{\E}{\mathbb{E}}

\DeclareMathOperator*{\argmax}{arg\,max}

\newtheorem{remark}{Remark}

\usepackage[accepted]{mlsys2024}

\mlsystitlerunning{SiDA-MoE: Sparsity-Inspired Data-Aware Serving for Efficient and Scalable  Large Mixture-of-Experts Models}

\begin{document}

\twocolumn[
\mlsystitle{SiDA-MoE: Sparsity-Inspired Data-Aware Serving for Efficient and Scalable Large Mixture-of-Experts Models}



\mlsyssetsymbol{equal}{*}

\begin{mlsysauthorlist}
\mlsysauthor{Zhixu Du}{duke}
\mlsysauthor{Shiyu Li}{duke}
\mlsysauthor{Yuhao Wu}{duke}
\mlsysauthor{Xiangyu Jiang}{clemson}
\mlsysauthor{Jingwei Sun}{duke}
\mlsysauthor{Qilin Zheng}{duke}
\mlsysauthor{Yongkai Wu}{clemson}
\mlsysauthor{Ang Li}{maryland}
\mlsysauthor{Hai "Helen" Li}{duke}
\mlsysauthor{Yiran Chen}{duke}
\end{mlsysauthorlist}

\mlsysaffiliation{duke}{Department of Electrical and Computer Engineering, Duke University, Durham, USA}
\mlsysaffiliation{maryland}{Department of Electrical and Computer Engineering, University of Maryland, College Park, USA}
\mlsysaffiliation{clemson}{Department of Electrical and Computer Engineering, Clemson University, Clemson, USA}

\mlsyscorrespondingauthor{Zhixu Du}{zhixu.du@duke.edu}

\mlsyskeywords{Machine Learning, MLSys}

\vskip 0.3in

\begin{abstract}
Mixture-of-Experts (MoE) has emerged as a favorable architecture in the era of large models due to its inherent advantage, i.e.,  enlarging model capacity without incurring notable computational overhead. Yet, the realization of such benefits often results in ineffective GPU memory utilization, as large portions of the model parameters remain dormant during inference. Moreover, the memory demands of large models consistently outpace the memory capacity of contemporary GPUs. Addressing this, we introduce SiDA-MoE (\textbf{S}parsity-\textbf{i}nspired \textbf{D}ata-\textbf{A}ware), an efficient inference approach tailored for large MoE models. SiDA-MoE judiciously exploits both the system's main memory, which is now abundant and readily scalable, and GPU memory by capitalizing on the inherent sparsity on expert activation in MoE models. By adopting a data-aware perspective, SiDA-MoE achieves enhanced model efficiency with a neglectable performance drop. Specifically, SiDA-MoE attains a remarkable speedup in MoE inference with up to $3.93\times$ throughput increasing, up to $72\%$ latency reduction, and up to $80\%$ GPU memory saving with down to $1\%$ performance drop. This work paves the way for scalable and efficient deployment of large MoE models, even with constrained resources. Code is available at: \href{url}{https://github.com/timlee0212/SiDA-MoE}.
\end{abstract}
]



\printAffiliationsAndNotice{} 

\section{Introduction}

\label{sec:intro}
Recently, rapid advances in large models with shocking performance have surprised the community in several areas, such as vision~\citep{ramesh2022hierarchical, kirillov2023segment, saharia2022photorealistic}, language~\citep{brown2020language, openai2023gpt4, smith2022using}, decision making~\citep{yang2023auto}, and robotics~\citep{vemprala2023chatgpt}. 
For example, GPT-4 has demonstrated its capability that is comparable or even exceeds human-level understanding on several tasks~\citep{openai2023gpt4}, and DALLE$\cdot$2 can generate astonishing high-quality images. The outstanding performance of large models heavily relies on the outrageous number of parameters, namely the scaling law~\citep{kaplan2020scaling}. Broadly speaking, the scaling law asserts that as the model size increases, various characteristics such as training loss, test performance, and the amount of required data exhibit predictable scaling behaviors. 

\begin{figure}[!h]
    \centering
    \includegraphics[width=0.9\columnwidth]{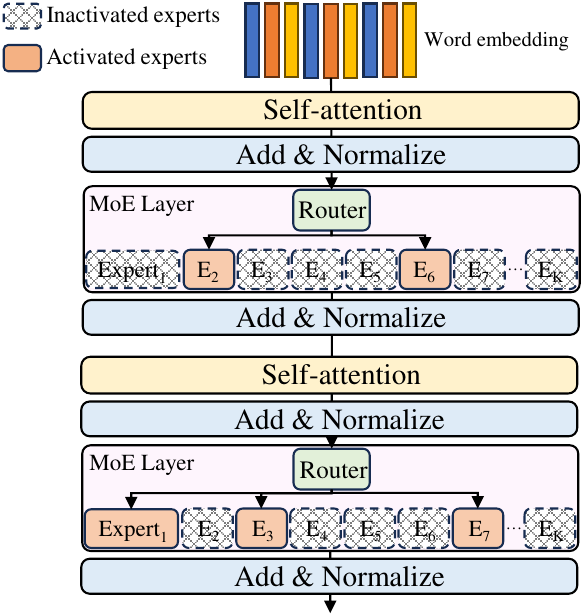}
    \caption{Diagram Showcasing the Architecture of MoE-based Transformers. Within each MoE layer only a limited number of experts are activated for inference.}
    \label{fig:architecture}
    \vspace{-1em}
\end{figure}

Mixture-of-Experts (MoE), a classical model architecture, enjoys the advantage that naturally fits the era of large models. 
MoE can improve the model's performance by drastically increasing the number of parameters while only incurring little computational overhead. 
Although the number of parameters involved in the forward pass of an MoE model remains almost unchanged, research \citep{fedus2022switch} suggests that augmenting parameter counts using the MoE architecture still conforms to the scaling law.
Encouraged by the advantage, many MoE-based large models have been proposed and achieved overwhelming performance in computer vision~\citep{li2023sparse, riquelme2021scaling, xue2022go}, natural language processing~\citep{shazeer2017, fedus2022switch}, 
Specifically, the Sparsely-Gated Mixture-of-Experts~\citep{shazeer2017} scales LSTM models to 137 billion parameters, which improves the model capacity by $1000\times$ with marginal computational overhead increase. Switch Transformers~\citep{fedus2022switch} scale to 1.6 trillion parameters with the same perplexity as T5-XXL~\citep{raffel2020exploring} while $4\times$ speedup during inference. 
However, the success of MoE comes with sacrifices in \textit{effective GPU memory} utilization, incurring large memory occupation while only a small fraction of parameters residing in the memory are effective for inference of the current batch. Fig.~\ref{fig:architecture} depicts the architecture of MoE-based transformers, where only a small portion of experts are activated in each MoE layer during each inference.

Further, with the trend of model scaling, we have observed a substantial gap between the memory demands of large models and the memory capacity of GPUs. For instance, in the past three years, the number of parameters in state-of-the-art models has scaled from 175 billion in GPT-3 \cite{brown2020language} to 1.76 trillion in the newly announced GPT-4 \cite{openai2023gpt4}, showing an over 10$\times$ increase. Contrarily, the memory capacity of high-end GPUs remains around 80GB \cite{choquette2023nvidia}, and commodity GPUs are still limited to 48GB or even smaller. This growing discrepancy motivates techniques to improve memory utilization efficiency. Thus, we seek to answer a compelling research question: 
\begin{center}
   \textit{How to serve large Mixture-of-Experts models in an efficient and scalable manner under constrained memory?}
\end{center}

Previous efforts have studied the efficiency problem of MoE models to some extent. Deepspeed-MoE \cite{rajbhandari2022deepspeed} optimizes the MoE module in the Deepspeed framework for efficient grouping and scheduling. A later version of the work \cite{aminabadi2022deepspeed} focused on optimizing the inference efficiency with optimized computation kernels and careful coordination of communication and parallelism. Tutel \cite{hwang2023tutel} enables adaptive parallelism and pipelining at runtime. However, these methods only focus on optimizing device-to-device communication but ignore the data-awareness,
not to mention exploiting the data-awareness to improve efficiency during inference. The data-awareness refers to a design where the technique or strategy is determined based on the incoming data. Our proposed framework embraces the data-awareness which brings three advantages. 
\textit{Firstly}, the data-awareness can squeeze the sparsity leading to a further increase in memory efficiency compared to previous methods. 
\textit{Secondly}, the data-awareness preserves the structure crucial for a sample's unique features, better maintaining the model's performance. 
\textit{Thirdly}, the data-awareness offers better adaptability since the framework varies according to data distribution.

\begin{table}
    \caption{\textit{Comparison of SiDA-MoE and Baseline Methods.} This table delineates the capabilities of various methods in terms of data-awareness, effective GPU memory utilization, and inference speed on large MoE models. SiDA-MoE excels in its data-aware approach with high effective GPU memory utilization and high inference speed on large MoE models.
    }
    \centering
    \vspace{0.3em}
    \resizebox{0.48\textwidth}{!}
    {
        \begin{tabular}{cccc}
        \toprule[1pt]
         Methods &  Data-aware & \makecell{Effective GPU \\ memory utilization} &\makecell{Inference speed \\on large MoE}\\ 
        \hline
        Standard  &  \ding{55} & low  & slow \\   
        Deepspeed  & \ding{55} & medium  & slow \\
        Tutel  & \ding{55}     & medium  & slow \\
        \hline
        \textbf{SiDA-MoE}  & \textbf{\ding{51}}      & \textbf{Extremely high} & \textbf{Extremely high}\\
        \toprule[1pt]
        \end{tabular}
    }
    \label{tab:summarization}
    \vspace{-0.6em}
\end{table}

In this paper, we present an efficient inference system, i.e., SiDA-MoE (\textbf{S}parsity-\textbf{i}nspired \textbf{D}ata-\textbf{A}ware),  for serving large MoE models. By noticing that modern server CPUs support terabytes (TB) of main memory, dwarfing GPU capacity, SiDA-MoE dynamically leverages both main memory and GPU memory by exploiting sparsity in MoE models in a data-aware manner. We summarize the comparison in Table~\ref{tab:summarization} between SiDA-MoE and baselines. Specifically, 
SiDA-MoE contains two threads that run in parallel, an inference thread and a hash-building thread. 
The hash-building thread exploits the sparsity of expert activation in a data-aware manner, whose core is a network-based hash function.
Specifically, the hash function is an offline trained predictor that predicts the experts to be activated. In this work, we employ a LSTM~\citep{hochreiter1997long} with sparse attention and a truncated knowledge distillation to boost the performance of the hash function.
The inference thread offloads inactivated experts predicted by the hash-building thread to maximize effective GPU memory utilization. Besides, SiDA-MoE also brings significant speedup during inference.
Our contributions are summarized as follows:
\begin{itemize}
\vspace{-2.5mm}
    \item To the best of our knowledge, {SiDA-MoE} is the first \textit{sparsity-inspired} \textit{data-aware} system serving for efficient and scalable inference on large MoE models. 
    \vspace{-1mm}
    \item We propose an offline training strategy to build a data-aware hash function deployed in SiDA-MoE that replaces the router function in MoE layers. Our design boosts the throughput of MoE models up to $3.93\times$ and reduces the latency down to $28\%$.
    \vspace{-1mm}
    \item Our offloading scheme achieves up to $80\%$ GPU memory saving with only less than 1\% performance drop. Our hash function can achieve up to $99\%$ prediction accuracy on expert activation. 
    \vspace{-1mm}
\end{itemize}

The paper is organized in the following manner: In Section 2, we will introduce the background and motivation. Section 3 is devoted to the framework of SiDA-MoE, our observation of the expert activation pattern, and our design of the offline trained hash function. In Section 4, we present our experimental results. Section 5 and 6 are devoted to related works and discussions, respectively. Finally, Section 7 will conclude the paper.

\section{Background and Motivation}
In this section, we present the background and motivation for SiDA-MoE. We employ the following notation throughout this paper: $a$ denotes a scalar, $\va$ represents a vector, $\mA$ signifies a matrix, and $\sA$ indicates a set. The notation $[K]$ is used to denote the set $\{1,2,\ldots,K\}$. Unless explicitly stated otherwise, all experiments are conducted with a batch size of 1 to isolate the influence of batch size. 

\label{sec:related_work}
\subsection{Mixture of Experts}
Since the first proposal of Mixture-of-Experts (MoE)~\cite{jacobs1991adaptive, jordan1994hierarchical}, different MoE models have been proposed based on various experts models, for example, hidden Markov models~\citep{jordan1996hidden}, Gaussian Process~\citep{tresp2000mixtures}, and support vector machine~\citep{collobert2001parallel}. With the rise of deep learning, Eigen et al. propose the use of several sets of routers and experts to build a stacked model, namely Deep MoE~\cite{eigen2013learning}. 

A MoE layer consists of a router function, denoted as $h(\cdot; \mW_r)$, followed by $K$ experts in parallel, denoted as $\{f_i(\cdot;\vtheta_i )\}_{i=1}^K$. Usually, the router function is set as a linear function, i.e.,  $h(\rvx; \mW_r) = {\mW_r}^\top\rvx$ where $\mW_r \in \sR^{d \times K}$ for input $\rvx \in \sR^d$, and experts are multi-layer perceptrons (MLPs) with a non-linear activation function~\citep{chen2022towards, fedus2022switch, shazeer2017}. The output of a MoE layer takes the form: 

\begin{equation}
    \label{eq:moe}
    M(\rvx;\mW_r, \vtheta_1, ..., \vtheta_K) = \sum_{i\in\sI} \alpha_i(\rvx)f_i(\rvx;\vtheta_i),
\end{equation}
where $\sI$ contains the selected indices of experts and the scaling factor $\alpha_i$ is defined as $$\alpha_i(\rvx)=\frac{\exp\{\mW_r[:,i]^\top\rvx\}}{\sum_{j=1}^K \exp\{\mW_r[:,j]^\top\rvx\}}.$$
Different selection mechanism of $\sI$ leads to different models. The soft-routing model~\citep{jordan1994hierarchical} selects all experts, i.e., $\sI = [K]$, which leads to high computational overheads. The switch-routing model~\citep{fedus2022switch} selects the top-$1$ expert, i.e., $\sI = \argmax_{i\in[K]} \alpha_i(\cdot)$, introducing little extra computational overhead.

\subsection{Low Effective Utilization of GPU Memory }
\begin{figure}[!h]
    \centering
    \includegraphics[width=0.9\columnwidth]{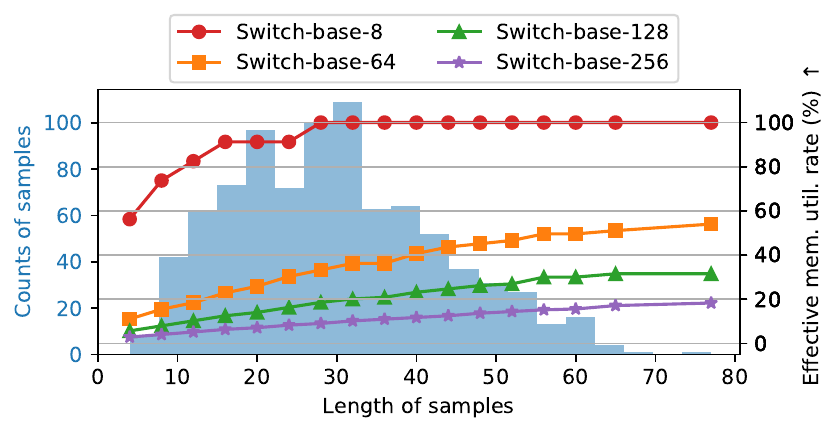}    
    \vspace{-9pt}\caption{Memory Efficiency of Switch Transformers on SST2. The $x$-axis represents the length of the sentence and the bar records the counts of sentences of corresponding length. The line represents the effective memory utilization for Switch Transformer on SST2 with a varied sentence length. Down to $5\%$ utilization can be observed for large models.} 
    \label{fig:glue_gpu_save}
\end{figure}

Encouraged by the advantage of MoE-based large models that drastically increasing the number of parameters leads to little computational overhead, many large-scale architectures have been proposed such as the Sparsely-Gated MoE~\citep{shazeer2017}, Gshard~\citep{lepikhin2020gshard}, and Switch Transformers~\citep{fedus2022switch}. Specifically, the Sparsely-Gated MoE proposes a trainable router function to determine the expert to be activated for each sample, which makes it possible to build very large MoE-based models as it improves the computational efficiency by a large margin compared to the soft-routing selecting all experts. 
The Sparsely-Gated MoE scales LSTM models to 137 billion parameters achieving outstanding performance. 
Switch Transformers, the most widely used transformer-based large MoE, converts T5 models~\citep{raffel2020exploring} to their MoE versions. All Switch Transformers outperform their foundation dense model with the same FLOPs.

In our study, we found that large MoE models do not efficiently utilize GPUs. As shown in Eq.~\ref{eq:moe}, we denote an expert as activated if $i \in \sI$. Inactivated experts remain idle in the forward pass, leading to low effective GPU memory utilization. \textit{Effective GPU memory} refers to the memory storing parameters that are effective for the forwarding of the model. The inactivated experts occupy a large amount of GPU memory while remaining idle, leading to low effective GPU memory utilization. To quantitatively analyze the GPU memory utilization, we provide a summary of Switch Transformers on model size and MoE layer size in Table~\ref{tab:model_size}. It is shown that for all Switch Transformers, especially the large ones, MoE layers occupy a large portion of GPU memory. Meanwhile, most of the parameters of the MoE layers are idle during one forward pass. To ascertain the amount of ineffective GPU memory, we feed samples from the SST2 dataset to Switch Transformers and record the corresponding effective memory utilization rates. The results are depicted in \cref{fig:glue_gpu_save}. For large Switch Transformers such as Switch-base-128 and Switch-base-256, the ineffective GPU memory for short sentences is around 24GB and 50GB, respectively. Even for the longest sentences with 80 tokens, the ineffective GPU memory is around 20GB and 46GB, respectively. 
Our method, SiDA-MoE, can save all ineffective GPU memory, outperforming baselines by a large margin. 
Further results on GPU memory reduction across datasets can be found in Section~\ref{sec:exp}.

\begin{table}
    \caption{\textit{Memory Occupation of Switch Transformers}. This table highlights the allocation of parameters in gigabytes (GB) for different models. MoE parameters dominate memory usage, especially in larger models. In contrast, mainstream GPUs peak at 48GB, with many at 24GB, while mobile GPUs range from 4GB to 12GB.
    }
    \centering
    \vspace{0.5em}
    \resizebox{0.48\textwidth}{!}
    {
        \begin{tabular}{cccc}
        \toprule[1pt]
         &  Model (GB) & MoE (GB) & Percentage ($\%$)\\ 
        \hline
        Switch-base-8  & 2.298 & 1.7932  & 78.03   \\   
        Switch-base-64  & 14.112 & 13.608    & 96.42 \\
        Switch-base-128  & 27.614 & 27.11 & 98.17 \\
        Switch-base-256  &  54.62 & 54.114 & 99.07 \\
        \toprule[1pt]
        \end{tabular}
    }
    \label{tab:model_size}
\end{table}

\subsection{High MoE Overhead}
\begin{figure}
    \centering
    \includegraphics[width=0.9\linewidth]{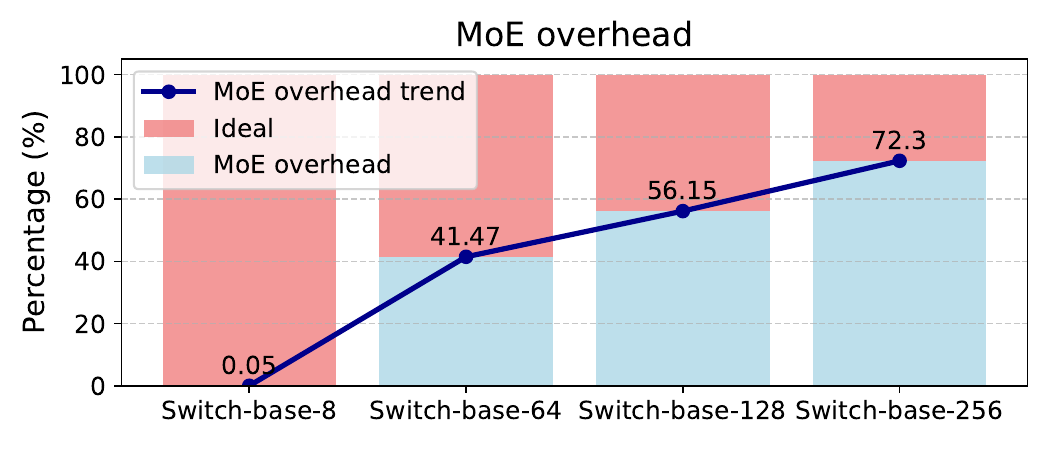}
    \vspace{-9pt}    \caption{MoE Overhead on SST2. The bar depicts the percentage breakdown for MoE overhead and Ideal Inference time. Up to $72\%$ time on Switch-base-256 are occupied by MoE overhead, including expert selection, expert invocation, and communication. Notably, the occupation of expert selection overhead scales up as model size increases. 
    }
    \label{fig:expert_selection_latency}
\end{figure}
Apart from the low effective GPU memory utilization, we also observed a high MoE overhead that includes expert selection, expert invocation, and additional communication costs when inference with MoE architectures. 
We conduct experiments on SST2 with multiple Switch Transformer models and provide the profiling results in \cref{fig:expert_selection_latency}. It is shown that the MoE overhead consumes over $72\%$ of the total inference time for Switch-base-256, which is a bottleneck of inference on MoE models. Notably, the overhead associated with MoE escalates with the scale of the model, further emphasizing the imperative of addressing the bottleneck in inference efficiency. We account for this for the default implementation that invokes every expert, irrespective of whether any tokens are assigned to it to align with hardware for efficient computation with a huge amount of requests. 
In our modified implementation used as a proxy for ideal, where we replace the router with a lookup table, we only invoke an expert if there is token assigned to that expert to suit the resource constraints scenario.

\begin{remark}
    Given that we simulate an inference scenario with a small batch size, the input to each expert is minimal. Consequently, the invocation overhead surpasses the computation itself — meaning the number of experts called dictates the overall inference time. 
\end{remark}

\subsection{Sparse Activation of Experts in Large MoE Models}

\label{sec:sparse_moe}
\begin{figure}
    \centering
    \includegraphics[width=0.9\linewidth]{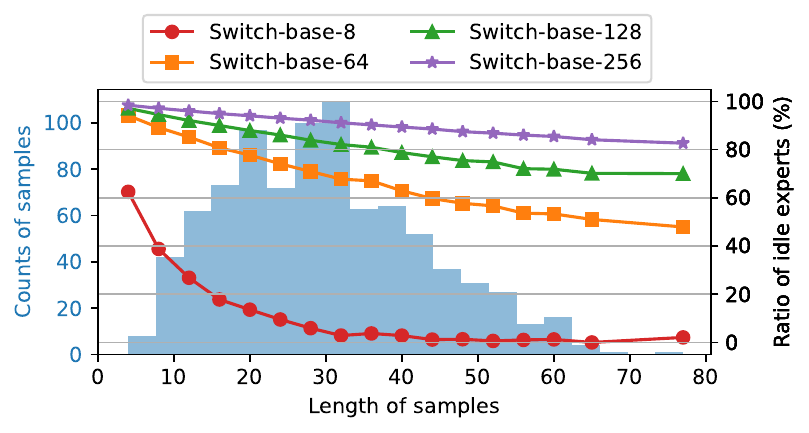}
    \vspace{-9pt}    \caption{Expert Activation in Switch Transformers on SST2. The $x$-axis denotes sentence length, with bars illustrating the counts of given lengths. The line depicts the ration of idle experts. Notably, Switch-base-256 and Switch-base-128 activate less than $20\%$ and $40\%$ of their experts, respectively. }
    \label{fig:sparsity}

\end{figure}

The sparse selection of experts is one of the critical observations that motivate SiDA-MoE. Our observation verifies that only a small portion of experts will be activated during inference.

For each token, the router function will select either top-$K$~\citep{shazeer2017} or top-$1$~\citep{fedus2022switch} experts inducing a token level expert activation sparsity. However, the sparsity on sentences, typically with 512 or 768 tokens, remains elusive. Not to mention in the training stage, an expert loading balance loss must be applied, which forces the router to assign an almost equal number of tokens to each expert. 
Otherwise, router's outputs will collapse to few experts leading to capacity degradation~\citep{chen2022towards}.

We test Switch Transformers with different number of experts on the SST2 dataset and report the sentence level sparsity in Fig.~\ref{fig:sparsity}. Our observation verifies that the sparse activation pattern still exists at the sentence level for large MoE models such as Switch-base-128 and Switch-base-256. As shown in the figure, down to less than $40\%$ experts and $20\%$ experts are activated for Switch-base-128 and Switch-base-256, respectively. Even for the longest sentences with around 80 tokens, the ratio of idle experts is still higher than $70\%$ for Switch-base-128 and $80\%$  for Switch-base-256. 

\begin{figure*}[ht]
    \centering
    \includegraphics[width=0.95\textwidth]{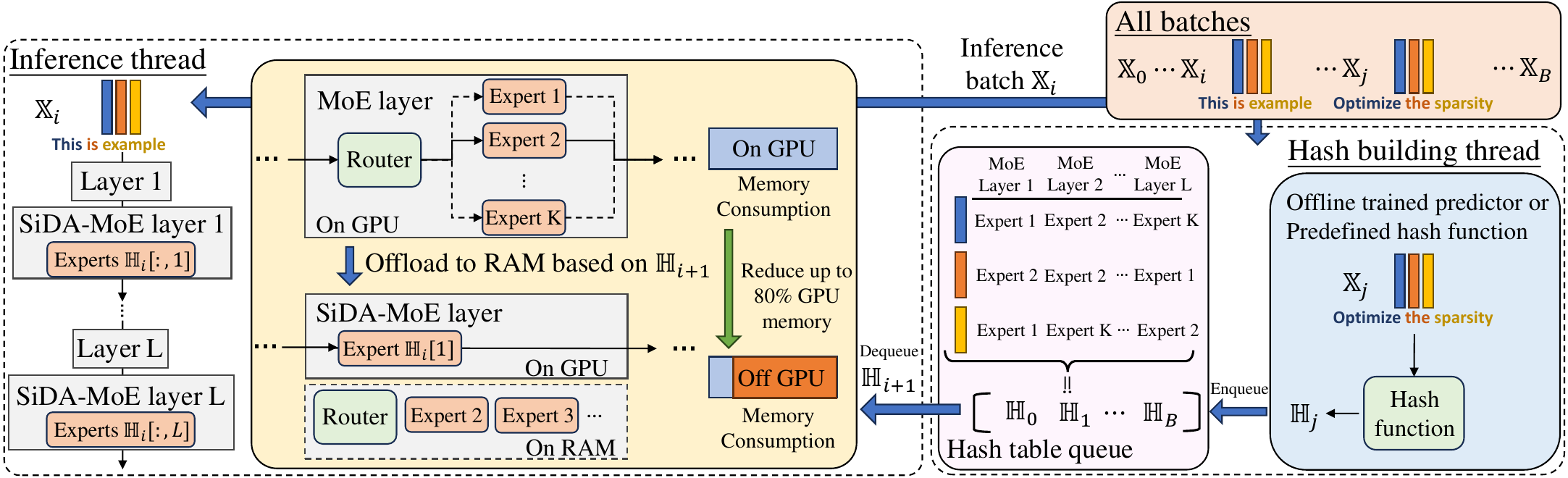}
    \caption{Overview of SiDA-MoE. SiDA-MoE contains two threads, the inference and hash-building thread, that run concurrently. As each batch $\sX_j$ arrives, the hash-building thread constructs the expert hash table \( \sH_j \) and queues it. In tandem, the inference thread processes the preceding batch \( \sX_i \), dynamically managing experts in MoE layers based on the hash table \( \sH_{i} \).}
    \label{fig:overview}
\end{figure*}
\section{SiDA-MoE} 
\label{sec:hash_moe}
\subsection{Overview: workflow}

We introduce a novel framework, Sparsity-inspired Data-Aware (SiDA-MoE), for efficient inference of large MoE models, whose overview is shown in Fig.~\ref{fig:overview}. SiDA-MoE contains two parallel threads that run simultaneously, namely the \textit{Inference thread} and the \textit{Hash-building thread}. 
Consider a sequence of incoming batches, batch $\sX_j$ is fed to the hash-building thread to build the hash table $\sH_j$ storing expert activation patterns for batch $\sX_j$, which will be pushed to the hash table queue. At the same time, the inference thread is handling the precedent batch $\sX_i$ and operating dynamical offloading on MoE layers based on the hash table $\sH_{i}$. 

\textbf{Hash-building thread.} The Hash-building thread consists of two components, a hash function and a hash table queue. For each incoming batch (\circled{1}-a), the hash function will determine experts to be activated for each token at each layer and the corresponding scaling factor $\alpha$ (\circled{1}-b). The predictions are stored in the hash table $\sH_j$ for the batch $\sX_j$ and pushed to the hash table queue (\circled{1}-c). 
The hash function can be a predefined hash function if the MoE model is trained with the Hash layer~\citep{roller2021hash}. More commonly, for the MoE model using trained router functions, such as Switch Transformers, the hash function will be offline trained. We propose hash function training techniques dedicated to modern MoE models,
which will be introduced in later sections.

\textbf{Inference thread.} The inference thread performs two tasks, i.e., dynamically load activated experts and offload inactivated experts according to the hash table built by the hash-building thread, and use the SiDA-MoE MoE layers to inference input batches. Specifically, for each incoming batch $\sX_{i}$ (\circled{2}-a), the inference thread will first pop the hash table $\sH_i$ from the hash table queue (\circled{2}-b) and remain idle if $\sH_i$ is not found. 
Notably, in practice, the inference thread takes a longer time to inference a batch than the hash-building thread to build a hash table for a batch. As a result, the inference thread never idles except at the very beginning. 
With the popped hash table $\sH_i$, the next step is to dynamically load and offload experts. Based on GPU memory budgets and the expert activation pattern of the current batch, the inference thread will load activated experts to GPU and offload inactivated experts to RAM (\circled{2}-c). A first-in-first-out (FIFO) scheme is applied on experts if no memory budgets remain. The dynamical loading task of a MoE layer will be done right after the finish of inference on the previous batch following the pipeline parallelism mechanism~\cite{huang2019gpipe}. Note that, in our system, all routers are offloaded to the main memory and do not participate in the forward pass. Lastly, the incoming batch $\sX_i$ will be forwarded using the SiDA-MoE MoE layers specific to $\sX_i$ (\circled{2}-d). Algorithm ~\ref{algo:end2end} shows the end-to-end pipeline.

\subsection{Design challenges}
In the design of SiDA-MoE, we spot three key challenges.

\textit{\textbf{Challenge 1: How to efficiently obtain experts that are to be offloaded beforehand?}} Given the observation that experts are activated sparsely, it is trivial to save GPU memory by offloading inactivated experts 
to RAM. However, this nai\"ve implementation sacrifices the latency since expert activation patterns are inaccessible without the output of the router functions. It incurs large overheads to move experts between CPU and GPU after each router function as it breaks the forwarding pipeline. We propose to use an offline-trained hash function to acquire the expert activation pattern before inference starts for each batch. Furthermore, we design the hash function to run independently of model inference and build a hash-building thread running in parallel with the inference thread to achieve the efficiency requirements. By employing the hash-building thread, SiDA-MoE achieves outstanding latency compared to baselines since the expert selection, dynamical offloading, and inference all run in parallel. 

\textit{\textbf{Challenge 2: How to leverage sparse cross-embedding dependency on experts activation to design a lightweight offline trained hash function?}} Considering the inference efficiency and the GPU memory consumption of the system, the hash function must be a lightweight predictor. However, simple predictors can hardly capture the contextual information of the sequence and can be easily distracted. Hence, it becomes crucial to enforce the predictor to focus on critical information. We empirically verify that there exists a sparse cross-embedding dependency on expert activation, i.e., a limited number of embeddings in the sequence jointly affect expert activation. This sparse cross-embedding dependency sheds light on the success of lightweight predictors. However, it is impractical and inefficient to rule out all possible outcomes to find the cross-embedding dependency for every token. In response to the challenge, we propose a sparse attention mechanism on LSTM that enforces the predictor to focus on the most important embedding automatically. 

\textit{\textbf{Challenge 3: How to improve the expert selection accuracy and approximate the scaling factor simultaneously?}}
The hash function needs to determine not only the expert activation but also the scaling factor $\alpha$ in Eq.~\ref{eq:moe}. As the scaling factor is derived from the SoftMax logits output from the model, it is natural to apply knowledge distillation (KD), setting the router functions as teacher models and the hash function as the student model. However, it is impossible for the hash function to approximate the scaling factor distribution over all experts by KD due to the limited capacity of the hash function. To solve this challenge, we propose to use a truncated knowledge distillation (TKD), where the KD loss is computed over the top-$T$ experts. However, the TKD cannot guarantee adequate prediction accuracy. We further add a cross-entropy loss to boost the prediction accuracy. 

We introduce how SiDA-MoE deals with each challenge in detail in the following sections.

\subsection{Data-Aware and Efficient Expert Activation Prediction}

SiDA-MoE proposes a data-aware solution to efficiently obtain the experts to be offloaded beforehand. Specifically, we propose to use an offline trained hash function that takes the sequence of embedding as input and predicts all the activated experts for each token in the sequence. Training data of the hash function are pairs of input token embeddings and MoE expert activation patterns.  SiDA-MoE, augmented by the data-aware expert activation prediction, enjoys two advantages while compromising little loss of model performance down to less than $1\%$. \textit{Firstly}, the system can acquire the activation pattern of each sample beforehand and operate dynamically loading and offloading according to the GPU memory budget without interrupting the inference process. \textit{Secondly}, since the hash function determines the expert activation across all the MoE layers for a sample independently of the inference, the system can build the hash function in a hash-building thread running in parallel with the inference thread. By doing this, we can remove the overhead caused by expert selection from the inference time, which boosts the throughput up to $3.93\times$. 

Previous works have also been proposed to improve the router function of MoE, such as the Hash layer~\citep{roller2021hash} and the Base layer~\citep{lewis2021base}.
SiDA-MoE is orthogonal to these router functions as they can be accommodated in the hash-building thread. 
For MoE models with trained routers, we propose to train an LSTM as the hash function with the sparse attention boosted with our truncated knowledge distillation, detailed in the following sections.

\begin{figure}[h!]
    \centering
    \includegraphics[width=0.85\linewidth]{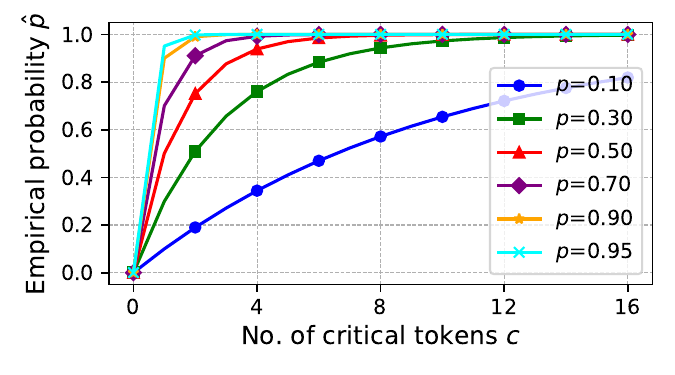}
    \vspace{-1.5em}
    \caption{Visualization of Eq.~\ref{eq:empirical_p} over Different $p$ and $c$.}
    \label{fig:sparsity_formula}
\end{figure}

\begin{figure}[h!]
    \centering
    \begin{subfigure}[b]{0.49\linewidth}
        \includegraphics[width=\linewidth]{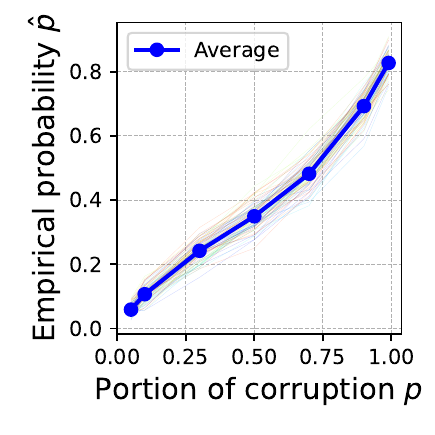}
        \vspace{-18pt}
        \caption{Tokens dependency.}
        \label{fig:embedding_dependency}
    \end{subfigure}
    \hfill
    \begin{subfigure}[b]{0.49\linewidth}
        \includegraphics[width=\linewidth]{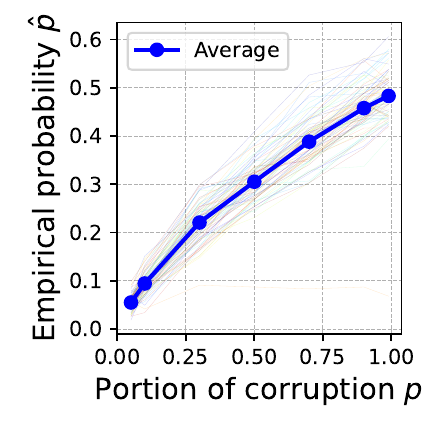}
        \vspace{-18pt}
        \caption{Positions dependency.}
        \label{fig:position_dependency}
    \end{subfigure}
    \vspace{-0.2em}
    \caption{Cross-embedding Dependency for Expert Activation on Switch-base-128 on C4. The $x$-axis shows the proportion of corruption, while the $y$-axis represents the empirical probability of expert activation change. Over 100 random embedding positions are examined, with the average trend displayed.}
    \label{fig:sparse_dependency}
\end{figure}
\subsection{LSTM with Sparse Attention}
\subsubsection{Sparse cross-embedding dependency on expert activation }
In the MoE layer, each word embedding will be fed to the router function to decide which expert to activate for inference of the token. However, the expert activation does not solely depend on the embedding corresponding to the token due to the self-attention layer before each MoE layer (shown in \cref{fig:architecture}), where the word embedding is mixed together. Because of the positional embedding, the position of tokens will also affect the expert activation. While the process by which embeddings collectively influence expert activation is complex, we identify a sparse cross-embedding dependency on expert activation, indicating that only a limited number of other tokens and positions are critical to the expert activation for the current token.

Suppose a sequence of length $L$, and let $c_i$ denote the number of critical tokens for the token at position $i$. We define the critical tokens as tokens in the sequence other than the selected $i$-th token, whose changes lead to a change in expert activation of the $i$-th token. In order to empirically verify that $c_i$ is a small number for all $i$, we consider finding a combinatorial equation involving $c_i$ and quantities we can measure. Consider selecting a set of tokens from the sequence excluding the $i$-th token, the probability that the set contains a critical token is formulated as below:
\begin{equation}
    \label{eq:empirical_p}
    \E[\hat{p}_i] = 1- \frac{\binom{L-1-c_i}{\lfloor pL \rfloor}}{\binom{L-1}{\lfloor pL \rfloor}}.
\end{equation}
where $\lfloor pL \rfloor$ denotes the size of the set and $p$ denotes the portion of selection over the sequence. Note that the probability that the selected set of tokens contains a critical token is equal to the probability that the $i$-th token's expert activation changes, denoted as $\hat{p}_i$, if we change all selected tokens in the set. We denote the process of changing the tokens in a sequence as `corruption.' Given Eq.~\ref{eq:empirical_p}, $p$ and $\hat{p}$ are quantities that we can empirically acquire, that is, by randomly selecting a portion $p$ of tokens, we can empirically measure the probability that the $i$-th token's expert activation changes. We show in \cref{fig:sparsity_formula} the relation between $c$ and $\hat{p}$ under different $p$.

Empirically, to study the token dependency of the token at position \(i\), the corruption is executed by randomly modifying a fraction \(p\) of chosen tokens from \( [L]-\{i\} \) to values distinct from their original and the \(i\)-th token. To examine the position dependency for the \(i\)-th token, the corruption also involves randomly choosing a fraction \(p\) of positions from \( [L]-\{i\} \) and swapping the token positions. 
We use the English division in the dataset C4~\citep{raffel2020exploring} to measure the probability that the $i$-th token's expert activation changes under different $p$, depicted in Fig.~\ref{fig:sparse_dependency}. We set the length $L=512$ and truncate or pad sentences which are not of length 512. 
We randomly test over 100 word embedding positions (i.e., 100  $i$'s) on Switch-base-128 and plot all of them in \cref{fig:sparse_dependency} with the average trend shown. \cref{fig:embedding_dependency} and \cref{fig:position_dependency} show the cross-embedding dependency of the token and position, respectively. Only a large portion of corruption leads to high chances of expert activation change, which demonstrates that most of the other tokens do not have an impact on the expert activation of the current token. 

By combining Fig.~\ref{fig:sparsity_formula} and Fig.~\ref{fig:sparse_dependency}, we can read the best approximation of $c_i$ based on different pairs of ($p$, $\hat{p}$) in Fig.~\ref{fig:sparse_dependency}, where we find that the best approximation of $\hat{c}$ ranges from $1$ to $4$ demonstrating the sparse cross-embedding dependency.

\subsubsection{Design of the hash function}
The design of the hash function must satisfy the following conditions: (1) be able to capture the sequential information, (2) be lightweight to preserve efficiency, and (3) be able to extract and focus on the critical embedding automatically. We adopt a 2-layer LSTM followed by a fully connected layer to align the first two conditions. Further, we add one fully connected layer to compress the embedding dimension. To achieve the third condition, we adopt the sparse attention mechanism with the SparseMax activation~\citep{martins2016softmax}. 

\textbf{Attention mechanism.}
The attention mechanism was first proposed in \cite{neural2015}, which has been proven to be influential in the realm of deep learning. The attention mechanism was proposed to allow the decoder to focus on different parts, resolving the problem that the encoder encodes the entire sentence. Given a query \(\vq\) and a set of key-value pairs \((\vk, \vv)\), the attention mechanism computes a weighted sum of values based on the similarity of the query to the keys. Formally, the attention weights \(\vw\) and the output \(\vo\) are computed as $\vo = \sum_i w_i \vv_i$ with
$$w_i = \frac{\exp(\text{score}(\vq, \vk_i))}{\sum_j \exp(\text{score}(\vq, \vk_j))},$$
where \(\text{score}(\vq, \vk)\) is a function that calculates the similarity between the query and a key. One common choice for \(\text{score}\) is the dot product of the query and key. 

We append one attention layer right after the LSTM layer where the key, value, and query are all set as the output sequence from LSTM. Consequently, each embedding will be a weighted sum of the sequence with weights proportional to the similarity between two vectors. The attention mechanism allows the predictor to pay different attention to different embeddings. However, the naive attention mechanism cannot impose a sparse focus. We further apply the SparseMax activation over $\vw$. 

\textbf{SparseMax activation.}
In contrast to the SoftMax activation, which provides a dense distribution, that is, non-zero probabilities assigned to all classes or positions, the SparseMax~\citep{martins2016softmax} provides a sparse distribution, where zero probability is assigned to many positions. We apply the SparseMax activation over the attention weights $\vw$ to obtain a sparse attention mechanism. 
Given an input vector \( \vw \in \mathbb{R}^L \), the SparseMax transformation is defined as:
\[ \text{SparseMax}(\vw) = \text{argmin}_{\vu \in \Delta^{L-1}} \left\| \vu - \vw \right\|_2^2, \]
where \( \Delta^{L-1} \) denotes the $(L-1)$-dimensional simplex, i.e.,
\[ \Delta^{L-1} = \{ \vu \in \mathbb{R}^L | \vu \geq 0, \sum_{i=1}^L u_i = 1 \}. \]
Although the expert selection is affected by other tokens in the sequence, the current token is always the most crucial on expert selection. Hence, we adopt the residual connection~\citep{he2016deep} to boost the performance right before the final fully connected layer.

\subsection{Truncated Knowledge Distillation}
The hash function of SiDA-MoE is required to predict the expert to be activated and the corresponding scaling factor $\alpha$. 
Knowledge distillation (KD)~\citep{hinton2015distilling}, which aims to minimize the distance of logits between the teacher and student model, should be the best training strategy for our hash function. However, the capacity of our hash function, 2-layer LSTM, is far less capable than the MoE model. The predictor cannot fully capture the behavior of logits of the router functions in the MoE model. The nai\"ve usage of KD greatly harms the performance of the system. 

We propose Truncated KD (TKD) to tackle the challenge. Different from the traditional KD, the truncated KD only considers positions with top-$T$ SoftMax logit, which helps the hash function focus more on predicting the scaling factor for experts with a higher chance of being activated. 
Notably, large $T$ can provide a smooth ground truth for the hash function, while small $T$ enforces the hash function to be more focused on fewer experts.
Further, we add the cross entropy loss to ensure the prediction accuracy. The training objective is $\lambda\mathcal{L}_{\text{CE}} + \mathcal{L}_{\text{TKD}}(T)$.

\begin{algorithm}[tb]
\caption{SiDA-MoE workflow}
\label{algo:end2end}
\small 
Assume batches of inputs $\{\sX_i\}_{i=0}^{B-1}$. Let $M$ denote the MoE model, $h$ denote the hash function, $\sH$ denote the hash table that stores the expert activation, and $k$ denote the index of experts.
\vspace{-1em}
\begin{multicols}{2}
\begin{algorithmic}
\STATE \textbf{Inference Thread:}
\FOR{each batch $\sX_i$}
    \STATE Dequeue $\sH_i$.
    \FOR{each MoE layer $l$ and each token $s$}
    \STATE Check Expert $\sH_i[l][s]$.
    \STATE  If not on GPU: to cuda.
    \ENDFOR
    \STATE $M_\text{SiDA-MoE} \shortleftarrow M$.
    \STATE output = $M_\text{SiDA-MoE}(\sX_i)$.
\ENDFOR
\STATE \textbf{end Inference Thread}
\end{algorithmic}

\columnbreak

\begin{algorithmic}
\STATE \textbf{Hash Building Thread:}
\FOR{each batch $\sX_i$}
    \FOR{each MoE layer $l$ and each token $s$}
    \STATE $\text{Expert } k \shortleftarrow h_l(\sX_i)[s]$.
    \STATE $\sH_i[l][s] \shortleftarrow \text{Expert } k$.
    \ENDFOR
    \STATE Enqueue $\sH_i$.
\ENDFOR
\STATE \textbf{end Hash Building Thread}
\end{algorithmic}
\end{multicols}
\vspace{-1em}
\end{algorithm}

\section{Experiment}
\label{sec:exp}
We extensively evaluate SiDA-MoE on different datasets. Specifically, we first show the GPU memory reduction ratio of SiDA-MoE demonstrating a memory saving up to $80\%$. We then report the throughput and latency of SiDA-MoE and baselines, where SiDA-MoE achieves up to $3.93\times$ improvements in terms of throughput with little performance degradation down to less than $1\%$. Our hash function achieves a prediction accuracy of up to $99\%$. Also, SiDA-MoE achieves the best efficiency under different GPU memory budgets.

\textbf{Implementation.} We implement the proposed SiDA-MoE framework atop the readily available Switch Transformer implementation in transformer \cite{wolf2019huggingface}, albeit not without substantial additional engineering effort. Enabling performant slice extraction poses challenges, as the MoE must maintain fine-grained associations between experts and hash table slices across layers and iterations. We optimize the parallel invocation of experts through meticulous inter-thread coordination, as naive parallelism introduces serious race conditions. The SiDA-MoE manager tackles intricate scheduling across the main training thread and the concurrent prediction thread, synchronizing via a shared queue that demands careful contention management. The main thread must then judiciously merge predictor outputs with the model state to orchestrate expert device placement, avoiding costly overheads like GPU-CPU data transfers.

\textbf{Setup.} We select three baselines namely, Standard, Deepspeed, and Tutel. The Standard baseline refers to the standard inference of the model. The Deepspeed refers to the Deepspeed implementation~\citep{aminabadi2022deepspeed} of the model, and the Tutel~\citep{hwang2023tutel} is designed for MoE models by enabling adaptive parallelism. 
We select three datasets from GLUE~\citep{wang2018glue} and SuperGLUE~\citep{wang2019superglue}. Specifically, we select SST2 and MRPC from GLUE for short sentences and mid-length sentences, and MultiRC from SuperGLUE for long sentences. 
We test most of the experiments on a server with an A-100 80GB GPU and 64 Intel(R) Xeon(R) Platinum 8358 CPU @ 2.60GHz CPUs. 
We investigate Switch-base-8, Switch-base-64, Swicth-base-128, and Switch-base-256 on efficiency, where the number indicates the number of experts in each MoE layer in the Switch Transformer. And we select Switch-base-8 and Switch-base-128 to fine-tune on selected datasets as the representatives on accuracy analysis, considering the representativeness and limited resources.
Our hash function in the hash building thread is trained on the train set of the dataset with the true hash table and evaluated on the test set of the dataset.

\textbf{Evaluation metrics.} We follow standard evaluation metrics for SST2, MRPC, and MultiRC, i.e., classification accuracy for SST2 and F1 score for MRPC and MultiRC. Further, we evaluate the fidelity of SiDA-MoE, which refers to how much performance can be preserved compared to original models. We refer the hash hits rate as the prediction accuracy on the expert activation of our hash function.

\textbf{Hyperparameters} We use AdamW~\citep{loshchilov2018decoupled} optimizer for fine-tuning the Switch Transformers and training the hash function. We set the batch size as 1 when measuring the latency and memory usage to eliminate the disturbance of the batch size. We select $T=30$ in the truncated KD with learning rate $5e-5$, batch size $64$, $\lambda=0.005$, and train to converge. For fine-tuning Switch Transformers, we set learning as $5e-5$ and fine-tune with $16000$ max steps.  We select top-$1$ experts from the hash function for SST2 and top-$3$ experts for MRPC and MultiRC when evaluating SiDA-MoE.

\subsection{GPU Memory Saving}
\begin{figure*}[h]
    \centering
    \includegraphics[width=\linewidth]{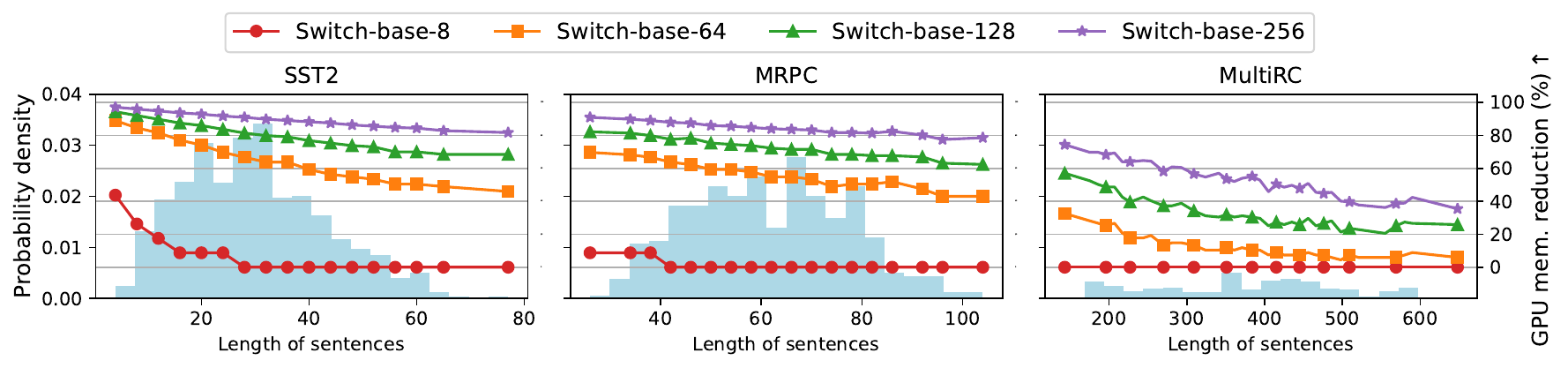}
    \vspace{-20pt}
    \caption{GPU Memory Reduction Rate by SiDA-MoE for Switch Transformers Across Datasets. SiDA-MoE achieves over $60\%$ and $80\%$ reduction on SST2 and MRPC for Switch-base-128 and Switch-base-256, respectively. And in MultiRC, with sentence lengths of 200-500, memory reductions of over $40\%$ for Switch-base-256 and $20\%$ for Switch-base-128 are noted.}
    \label{fig:gpu_mem_save}
\end{figure*}

We report the GPU memory saving in \cref{fig:gpu_mem_save}. For short sentences in SST2, SiDA-MoE can achieve over $80\%$ GPU memory reduction. For samples in MRPC whose lengths are clustered between 50 and 80, the GPU memory reduction remains substantial, yielding savings of 6.28GB and 19.84GB GPU memory for Switch-base-128 and Switch-base-256, respectively. Furthermore, even when processing long paragraphs in MultiRC with lengths ranging from 200 to 500, the rate of GPU memory reduction retains over $40\%$ and $20\%$, leading to a save of 4.52GB for Switch-base-128  and 9.92GB for Switch-base-256. 

\subsection{Latency and Throughput}
\begin{figure*}[h]
    \centering
    \includegraphics[width=\linewidth]{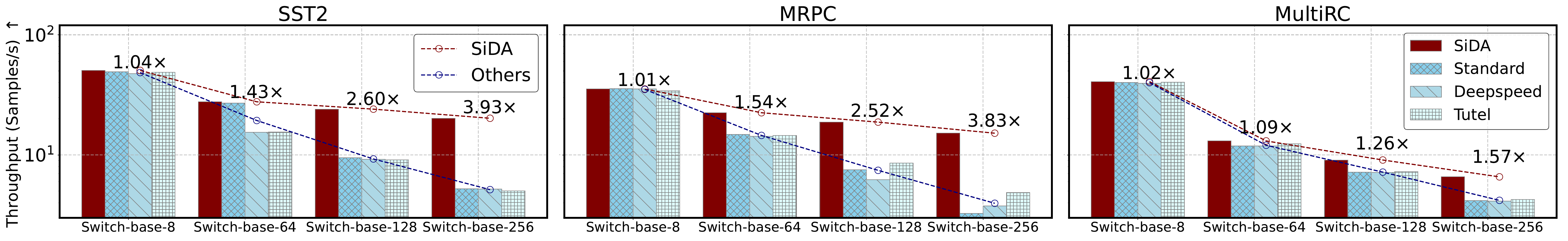}
    \vspace{-18pt}
    \caption{Throughput of Different Methods for Switch Transformers Across Datasets. SiDA-MoE achieves outstanding throughput for large MoE models on all three datasets with various sentence length and comparable results for small MoE models. Specifically, SiDA-MoE achieves $2.60\times$, $3.93\times$ more throughput on SST2, $2.52\times$, $3.83\times$ more on MRPC, and $1.26\times$, $1.57\times$ more throughput on MultiRC for Switch-base-128 and Switch-base-256, respectively.}
    \label{fig:throughput}
\end{figure*}

\begin{figure*}[h]
    \centering
    \includegraphics[width=\linewidth]{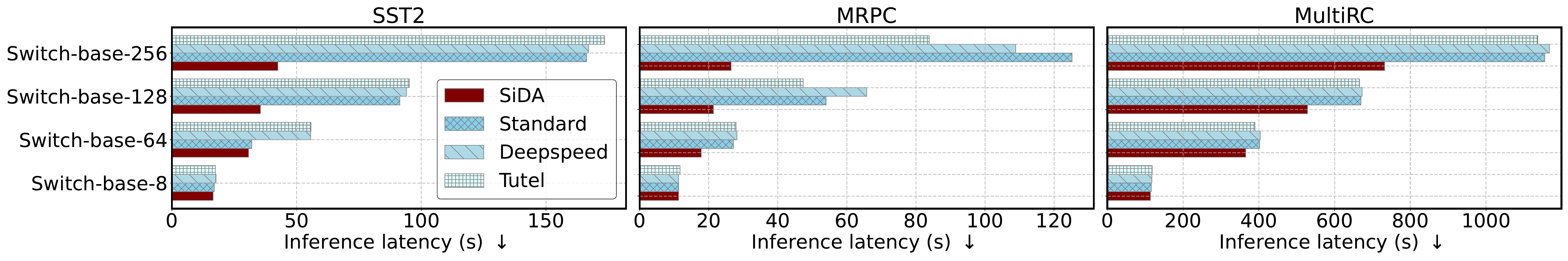}
    \vspace{-20pt} \caption{Comparison of Inference Latency Across Different Methods. SiDA-MoE consistently outperforms baselines, especially evident on Switch-base-256 model with latency reduced down to $28\%$. Notably, improvements are more pronounced as sentence lengths decrease. 
    }

    \label{fig:latency}
\end{figure*}
Apart from the GPU memory saving, SiDA-MoE also achieves overwhelming efficiency in terms of throughput and latency (see Fig.~\ref{fig:throughput}). Specifically, SiDA-MoE exceeds the average of baselines by $2.60\times$ and $3.93\times$ on throughput for large MoE models such as Swicth-base-128 and Switch-base-256 on SST2. Even for MultiRC containing long sentences, SiDA-MoE exceeds the average throughput of baselines by $1.26\times$ on Switch-base-128 and $1.57\times$ on Switch-base-256. 

We also investigate the inference latency of SiDA-MoE and baselines (see Fig.~\ref{fig:latency}). For large MOE models such as Switch-base-128 and Switch-base-256, SiDA-MoE reduces the inference latency to $25\%$ on SST2 and MRPC and to $60\%$ on MultiRC. The improvements come from our design of the hash-building thread that resolves the MoE overhead.

\subsection{Efficiency under Limited GPU Memory Budgets}
\begin{figure}[t]
    \centering
    \includegraphics[width=0.9\linewidth]{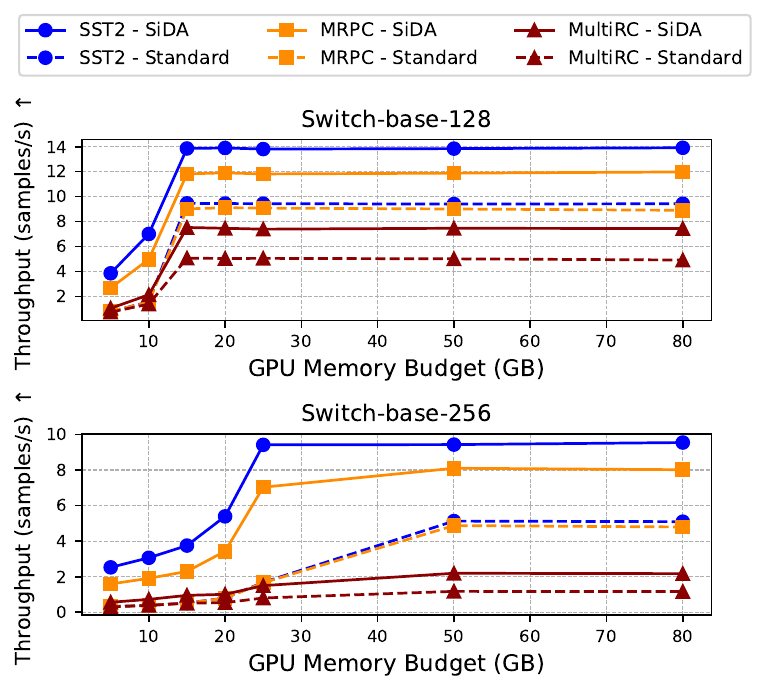}
    \vspace{-9pt}
    \caption{Throughput Efficiency Relative to GPU Memory Budget. SiDA-MoE's advantage is particularly pronounced in constrained GPU memory scenarios, showcasing its superior efficiency by offloading experts compared to the conventional model parallelism, here denoted as 'Standard'. }
    \label{fig:mem_lat_tradeoff}
    \vspace{-1em}
\end{figure}

We investigate the efficiency under different GPU memory budgets with different offloading methods on Switch-base-128 and Switch-base-256 since large MoE models are more resource-sensitive. Under a limited GPU memory budget, SiDA-MoE will offload and cache inactivated experts in a first-in-first-out manner\footnote{For fair comparison with baselines, we use FIFO, although other strategies could also be effective.}, while all other baselines implement the model parallelism, where only layers required for inference will be kept on the GPU. The results of throughput versus GPU memory budgets are shown in Fig.~\ref{fig:mem_lat_tradeoff}. SiDA-MoE achieves better throughput under all GPU memory budgets across all datasets, demonstrating that SiDA-MoE employs a better offloading strategy under limited GPU memory budgets.

\subsection{Capabilities as Pretrained Models}
\begin{table}[t]
    \vspace{-0.5em}
    \caption{Evaluation of SiDA-MoE's Perplexity. SiDA-MoE retains the perplexity well, especially for large Switch Transformers. For example, the perplexity only drops $3.52$ for Switch-base-256. 
    }
    \centering
    \vspace{0.5em}
    \resizebox{0.42\textwidth}{!}
    {
        \begin{tabular}{lcc}
        \toprule[1pt]
        Backbone         & Pretrained ppl. ($\downarrow$) & SiDA-MoE ppl. ($\downarrow$)\\ 
        \hline
        Switch-base-8    & $6.68$ & $18.49$\\
        Switch-base-64   & $4.93$ & $11.84$\\ 
        Switch-base-128  & $4.86$ & $11.73$\\
        Switch-base-256  & $4.59$ & $8.11$\\
        \toprule[1pt]
        \end{tabular}
    }
    \label{tab:perplexity}
\end{table}

We replace the router function with its approximation. Theoretically, our hash function approximates the density distribution of the router function, so the output will be similar. Empirically, to provide better sense on the performance degradation of Switch Transformers as pretrained models, we measure and report the perplexity of several Switch Transformers on the C4 dataset~\citep{raffel2020exploring}, where Switch Transformers are pretrained. Results are shown in \cref{tab:perplexity}. Note that there is a drop in perplexity degradation as models getting larger, since larger models have stronger resistance to experts miss-classification caused by the hash function. The following two sections show the results on finetuned downstream tasks.

\subsection{Fidelity Analysis}

\begin{table}[t]
    \vspace{-0.6em}
    \caption{{Evaluation of SiDA-MoE's Performance Preservation. SiDA-MoE retains as much as $99\%$ of the performance on the Switch-base-8 model and maintains over $95\%$ on the Switch-base-128 model, resulting in down to less than $1\%$ performance drop. }
    }
    \centering
    \vspace{0.5em}
    \resizebox{0.46\textwidth}{!}
    {
        \begin{tabular}{lcccc}
        \toprule[1pt]
        Backbone&         & SST2 & MRPC & MultiRC \\ 
        \hline
        \multirow{3}{*}{Switch-base-8} &Finetuned    & $92.20$ & $89.14$ & $56.70$ \\   
                                       &SiDA-MoE  & $90.59$ & $86.91$ & $56.11$  \\
                                       &Fidelity & $98.25\%$ & $97.49\%$ & $98.95\%$  \\
        \hline
        \multirow{3}{*}{Switch-base-128} &Finetuned    & $93.57$ & $89.66$ & $59.95$\\   
                                         &SiDA-MoE  & $87.04$ & $83.01$ & $55.49$  \\
                                         &Fidelity & $93.02\%$ & $92.59\%$ & $92.56\%$  \\
        \toprule[1pt]
        \end{tabular}
    }
    \label{tab:accuracy_preserving}
\end{table}
We conduct the fidelity analysis to check how much performance SiDA-MoE can preserve. As Table.~\ref{tab:accuracy_preserving} shows,
SiDA-MoE can preserve up to nearly $99\%$ accuracy leading to a performance degradation down to less than $1\%$ for Switch-base-8. For Switch-base-128, the fidelity is up to $96\%$ leading to a performance loss down to $3\%$. Our results demonstrate the superiority of SiDA-MoE, which achieves low inference latency and low GPU memory occupation with negligible loss on the model's performance.

\subsection{Hash Hits Rate}
\begin{table}[t]
    \vspace{-0.5em}
    \caption{Top-3 Hash Hits Rate. Demonstrating SiDA-MoE's exemplary accuracy on expert activation prediction up to over $99\%$ across various models.
    }
    \centering
    \vspace{0.5em}
    \resizebox{0.37\textwidth}{!}
    {
        \begin{tabular}{lccc}
        \toprule[1pt]
        Backbone         & SST2 & MRPC & MultiRC \\ 
        \hline
        Switch-base-8    & $99.00\%$ & $97.41\%$ & $91.74\%$ \\   
        Switch-base-128  & $98.78\%$ & $98.65\%$ & $90.49\%$  \\
        \toprule[1pt]
        \end{tabular}
    }
    \label{tab:missing_hash_rate}
\end{table}

SiDA-MoE adopts a predictor to predict the experts to be activated for each token. We investigate the accuracy of the predictor in the hash-building thread, which we refer to as the hash hits rate. Results can be found in Table~\ref{tab:missing_hash_rate} where we report top-$3$ accuracy. For very long sentences, such as the MultiRC dataset, the hash hits rate can achieve over $90\%$.

\section{Related Work}
With the rise of LLM, efficient serving for large models has become a hot topic. Much research has been done by adopting classical model compression methods, such as knowledge distillation~\citep{fu2023specializing, li2023symbolic, tan2023industry, wang2023scott, wu2023lamini, gu2023knowledge, zhou2023distillspec, yuan2023distilling}, quantization~\citep{chee2023quip, frantar2022gptq, lin2023awq, cheng2023optimize, liu2023qllm, liu2023llm, shang2023pb, shao2023omniquant, xiao2023smoothquant, yuan2023rptq}, and pruning~\citep{frantar2023sparsegpt, ji2023pruning, ma2023llm, sun2023simple, xia2023flash, li2023losparse}. Further, others have been exploring more efficient network architectures~\citep{del2023skipdecode, liu2023deja, miao2023specinfer, jiang2023recyclegpt, ning2023skeleton, spector2023accelerating, xu2023tensorgpt}. Besides, some have tackled the efficiency problem from a data perspective by performing text compression~\citep{chevalier2023adapting, ge2023context, valmeekam2023llmzip, jiang2023llmlingua}. However, these works are not specifically designed for MoE models and ignore the sparse expert activation patterns. SiDA-MoE exploits the expert activation patterns to achieve efficient inference. Furthermore, SiDA-MoE is orthogonal to methods such as quantization and pruning, which can be applied to the activated expert networks.  

Deploying deep neural networks under resource-limited scenarios has long been an active research area~\citep{mao2017modnn, zhou2019edge, wang2020convergence, qiu2024deep}. The efficient deployment of MoE-based models has gain more and more attention recently. SE-MoE~\citep{shen2022se} considers a sophisticated communication scheduling strategy design to improve the throughput in inference. Specifically, SE-MoE uses dynamic scheduling to overlap movement from CPU memory and inference computation in GPU memory. However, we propose a look-ahead strategy in a data-aware manner. Further, we observe and exploit the inherited sparsity on expert activation in MoE, while SE-MoE considers distillation and compression to reduce the MoE graph. Besides, $\text{M}^\text{3}$ViT~\citep{fan2022m3vit} proposes a co-design framework and Edge-MoE~\citep{sarkar2023edge} accomplish the hardware design of  $\text{M}^\text{3}$ViT. Fan et al. introduce MoE-based models to multi-task learning to allow efficient on-device multi-task learning for both training and inference. SiDA-MoE is different in the sense that the sparse expert path in $\text{M}^\text{3}$ViT is enforced by the multi-tasks, while we observe and exploit the inherited sparse expert activation pattern in MoE based-model. 

We also notice several concurrent works specifically designed for efficient MoE-based model inference~\citep{huang2023towards, kong2023serving, yi2023edgemoe}. However, SiDA-MoE is orthogonal to these works, which focus on designing better scheduling for caching experts. SiDA-MoE explores a data-aware path that predicts the experts to be activated. The data-aware approach and the caching scheduling can be combined to achieve better efficiency. 

\section{Discussion}
\textbf{Enhanced Hierarchical Offloading.} While SiDA-MoE offers offloading capabilities between main memory and GPU memory, its limitations are defined by the storage capacity of the main memory. This poses challenges, especially when deploying massive models like Switch-c-2048 with almost 5TB of parameters. A logical progression would be to introduce a layered offloading mechanism that fluidly transfers experts between GPU memory, main memory, and SSD storage. Such an advanced hierarchical approach in SiDA-MoE would make it adept at handling models of any magnitude.

\textbf{Optimized Hash Graph for Expert Activation Storage.} Currently, SiDA-MoE utilizes an LSTM model to function as its hash system. It's evident that the expert activation is conditionally contingent upon the activation patterns observed in preceding MoE layers. To enhance efficiency, an ideal hash function could be designed as a graph. This graph would capture and store these conditional dependencies, enabling rapid and effective extraction of expert activation.

\section{Conclusion}
In summary, this paper presents SiDA-MoE, a novel data-aware method that adeptly addresses the challenges posed by the memory constraints of GPUs when serving expansive models, specifically leveraging the sparsity inherent in MoE architectures. Further, SiDA-MoE deploys an offline trained hash function running in the hash-building thread, which alleviates the expert selection overhead by a large margin. 
Through judicious utilization of both main and GPU memory, SiDA-MoE offers a promising route for serving large MoE models under limited GPU budgets with nearly zero performance setbacks.

\section{Acknowledgements}
This work was supported in part by NSF 2112562 and ARO W911NF-23-2-0224.




\clearpage
\bibliography{reference}
\bibliographystyle{mlsys2024}

\newpage

%


\end{document}